\documentclass[10pt,twocolumn,letterpaper]{article}

\usepackage{cvpr}
\usepackage{times}
\usepackage{epsfig}
\usepackage{graphicx}
\usepackage{amsmath,amsfonts,amsthm,amssymb}
\usepackage{color}

\usepackage[flushleft]{threeparttable}
\usepackage{array}
\usepackage{booktabs}
\usepackage{float}
\usepackage{multirow}

\newcommand{\myparagraph}[1]{\vspace{-1em} \paragraph{#1}}

\usepackage[pagebackref=true,breaklinks=true,letterpaper=true,colorlinks,bookmarks=false]{hyperref}

\cvprfinalcopy 

\def\cvprPaperID{4578} 

\begin{document}

\title{Overcoming Classifier Imbalance for Long-tail Object Detection \\ with Balanced Group Softmax}

\author{Yu Li$^{1,2,3}$\thanks{This work was supported by the National Key Research and Development Program of China (2017YFC0820605), and the National Natural Science Foundation of China (61525206, 61572472, U1703261, 61871004), and 242 project (2019A010). Jiashi Feng was partially supported by AI.SG R-263-000-D97-490, NUS ECRA R-263-000-C87-133 and MOE Tier-II R-263-000-D17-112. Yu Li was partially supported by the program of China Scholarships Council (No.201904910801).$\dag$: corresponding author.}, Tao Wang$^{3,4}$, Bingyi Kang$^{3}$, $^{\dag}$Sheng Tang$^{1,2}$, Chunfeng Wang$^{2}$, Jintao Li$^{1,2}$, Jiashi Feng$^{3}$ \\
\small$^1$Key Laboratory of Intelligent Information Processing, Institute of Computing Technology, Chinese Academy of Sciences, Beijing, China\\
\small$^2$University of Chinese Academy of Sciences, Beijing, China \\
\small$^3$Department of Electrical and Computer Engineering, National University of Singapore, Singapore \\
\small$^4$Institute of Data Science, National University of Singapore, Singapore \\
\tt\small{\{liyu,ts,jtli\}@ict.ac.cn,twangnh@gmail.com,kang@u.nus.edu,} \\ 
\tt\small{wangchunfeng14@mails.ucas.ac.cn,elefjia@nus.edu.sg}
}

\maketitle
\thispagestyle{empty}
\begin{abstract}

Solving long-tail large vocabulary object detection with deep learning based models is a challenging and demanding task, which is however under-explored.
In this work, we provide the first systematic analysis on the underperformance of state-of-the-art models in front of long-tail distribution. 
We find existing detection methods are unable to model few-shot classes when the dataset is extremely skewed, which can result in classifier imbalance in terms of parameter magnitude. Directly adapting long-tail classification models to detection frameworks can not solve this problem due to the intrinsic difference between detection and classification.
In this work, we propose a novel balanced group softmax (BAGS) module for balancing the classifiers within the detection frameworks through group-wise training. It implicitly modulates the training process for the head and tail classes and ensures they are both sufficiently trained,  without requiring any extra sampling for the instances from the tail classes.
Extensive experiments on the very recent long-tail large vocabulary object recognition benchmark LVIS show that our proposed BAGS significantly  improves the performance of detectors with various backbones and frameworks on both object detection and instance segmentation. It beats all state-of-the-art methods transferred from long-tail image classification and establishes new state-of-the-art.
Code is available at \url{https://github.com/FishYuLi/BalancedGroupSoftmax}.
\end{abstract}

\vspace*{-15pt}
\section{Introduction}
\vspace*{-5pt}
\label{sec:intro}
Object detection~\cite{ren2015faster,redmon2018yolov3, liu2016ssd, lin2017focal, law2018cornernet,cai2018cascade} is one of the most fundamental and challenging tasks in computer vision.
Recent advances are mainly driven by large-scale datasets that are manually balanced, such as PASCAL VOC~\cite{everingham2010pascal} and COCO~\cite{lin2014microsoft}. 
However in reality, the distribution of object categories is typically long-tailed~\cite{reed2001pareto}.  Effective solutions  that adapt state-of-the-art detection models to such class-imbalanced distribution are highly desired yet still absent. 
Recently, a long-tail large vocabulary object recognition dataset LVIS~\cite{gupta2019lvis} is released, which greatly facilitates object detection research in much more realistic scenarios.

\begin{figure}
\centering
\vspace*{-10pt}
\includegraphics[width=0.85\linewidth]{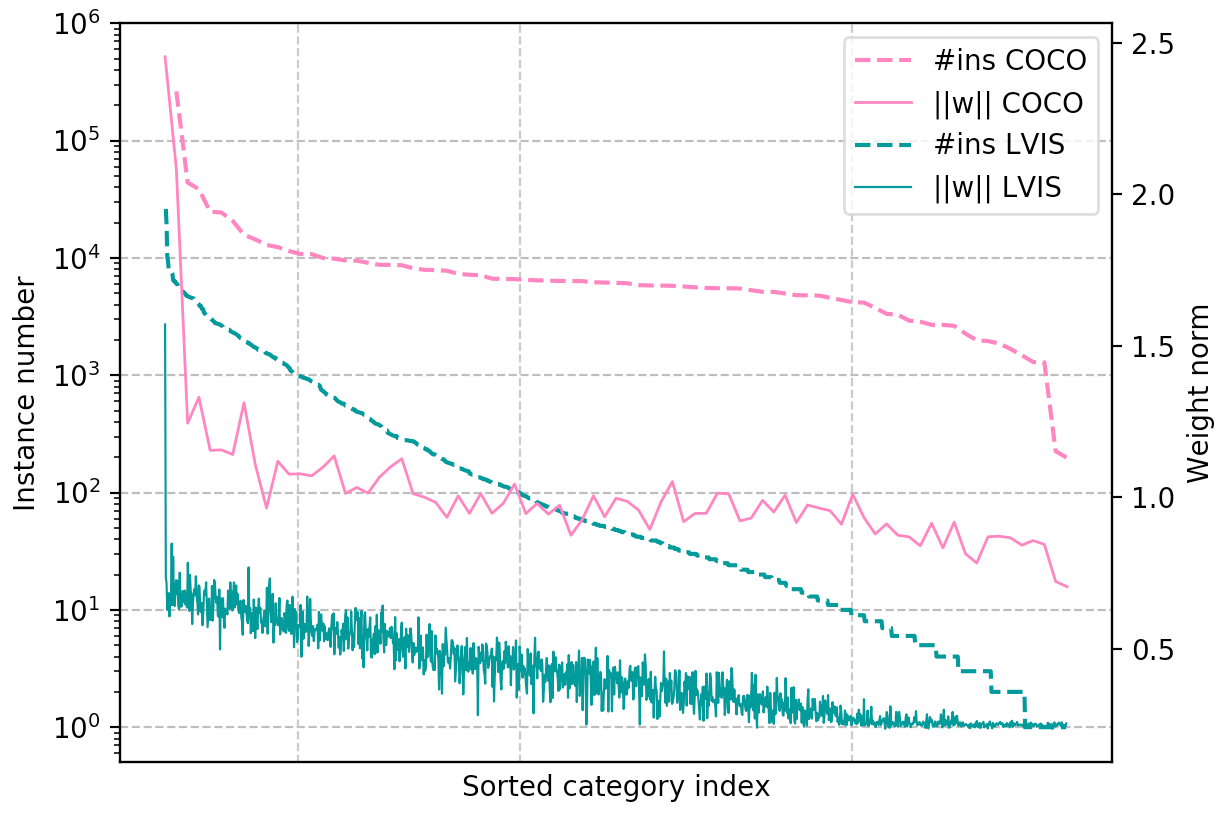}
	\vspace*{-6pt}
\caption{Sorted number of training instances (\#ins) for categories in COCO and LVIS training set, and the corresponding classifier weight norm $\Vert w \Vert$ from Faster R-CNN model trained on COCO and LVIS. The x-axis represents the sorted category index of COCO and LVIS. We align 80 classes of COCO with 1230 classes of LVIS for better visualization. Category 0 indicates background.}
\vspace*{-10pt}
	\label{fig:motivation}
\end{figure}

A straightforward solution to long-tail object detection is to train a well-established detection model (\eg, Faster R-CNN~\cite{ren2015faster}) on the long-tail training data directly.
However, big performance drop would be observed when adapting detectors designed for fairly balanced datasets (\eg, COCO) to a long-tail one (\eg, LVIS), for which the reasons still remain unclear due to multiple entangled factors.
Inspired  by~\cite{kang2019decoupling}, we decouple the representation and classification modules within the detection framework, and find the weight norms of the proposal classifier corresponding to different categories are severely imbalanced, since low-shot categories get few chances to be activated.
Through our analysis, this is one direct cause of the poor long-tail detection performance, which is intrinsically induced by data imbalance.
As shown in Figure~\ref{fig:motivation}, we sort the category-wise classifier weight norms of models trained on COCO and LVIS respectively by the number of  instances in the training set.
For COCO, the relatively balanced data distribution leads to relatively balanced weight norms for all categories, except for \emph{background} class (CID=0, CID for Category ID).
For LVIS, it is obvious that the category weight norms are imbalanced and positively correlated with the number of training instances.
Such imbalanced classifiers (w.r.t. their parameter norm) would make the classification scores for low-shot categories (tail classes)  much smaller than those of many-shot categories (head classes).
After standard softmax, such imbalance would be further magnified thus the classifier wrongly suppresses the proposals predicted as low-shot categories.

The classifier imbalance roots in data distribution imbalance\textemdash classifiers for the many-shot categories would see more and diverse training instances, leading to dominating magnitude. One may consider using solutions to long-tail classification to overcome such an issue, 
including re-sampling training instances to balance the distribution~\cite{han2005borderline, drummond2003c4, shen2016relay, mahajan2018exploring} and re-weighting classification loss at category level~\cite{cui2019class, cao2019learning, huang2019deep} or instance level~\cite{lin2017focal, shu2019meta}.
The re-sampling based solutions are applicable to detection frameworks, but may lead to increased training time and over-fitting risk to the tail classes.
Re-weighting based methods are unfortunately very sensitive to hyper-parameter choices and not well applicable to detection frameworks due to difficulty in dealing with the special \emph{background} class, an extremely many-shot category. We empirically find none of these methods works well on long-tail detection problem.

In this work, to address the classifier imbalance, we introduce a simple yet effective balanced group softmax (BAGS) module into the classification head of a detection framework. 
We propose to put object categories with similar numbers of training instances into the same group and compute group-wise softmax cross-entropy loss separately. Treating categories with different instance numbers separately can effectively alleviate the domination of the head classes over tail classes. However, due to the lack of diverse negative examples for each group training, the resultant model suffers too many false positives. Thus, BAGS further adds a category \emph{others} into each group and introduces the \emph{background} category as an individual group, which can alleviate the suppression from head classes over tail classes by keeping their classifiers balanced while preventing false positives by categories \emph{background} and \emph{others}.

We experimentally find BAGS works very well.
It improves by 9\%\ -- 19\% the performance on tail classes of various frameworks including Faster R-CNN~\cite{ren2015faster}, Cascade R-CNN~\cite{cai2018cascade}, Mask R-CNN~\cite{he2017mask} and HTC~\cite{chen2019hybrid} with ResNet-50-FPN~\cite{he2016deep, lin2017feature} and ResNeXt-101-x64x4d-FPN~\cite{xie2017aggregated} backbones consistently on the long-tail object recognition benchmark LVIS~\cite{gupta2019lvis}, with the overall mAP lifted by around 3\% -- 6\%.

To sum up, this work makes following contributions:
\begin{itemize}
\vspace{-0.5em}
 \item{Through comprehensive analysis, we reveal the reason why existing models perform not well for long-tail detection, \textit{i.e.} their classifiers are imbalanced and not trained equally well, reflected by the observed imbalanced classifier weight norms.}
\vspace{-0.5em}
 \item{We propose a simple yet effective balanced group softmax module to address the problem. It can be easily combined with object detection and instance segmentation frameworks to improve their long-tail recognition performance.}
\vspace{-0.5em}
 \item{We conduct extensive evaluations with state-of-the-art long-tail classification methods for object detection. Such benchmarking not only deepens our understandings of these methods as well as the unique challenges of long-tail detection,  but also provides reliable and strong baselines for future research in this direction.}
\end{itemize}

\section{Related Works}

Compared with balanced distribution targeted object detection~\cite{girshick2015fast, ren2015faster, cai2018cascade}, and few-shot object detection~\cite{kang2019few, chen2018lstd, zhang2019canet, fan2019few}, the challenging and practical long-tail object detection problem is still underexplored.
Though Ouyang \etal~\cite{ouyang2016factors} proposes the concept of long-tail object detection, their work focuses on the imbalanced training data distribution on ILSVRC DET dataset~\cite{deng2009imagenet} without few-shot setting for tail classes like LVIS~\cite{gupta2019lvis}.
\cite{gupta2019lvis} proposes repeat factor sampling (RFS) serving as a baseline.
Classification calibration~\cite{wang2019classification} enhances RFS by calibrating classification scores of tail classes with another head trained with ROI level class-balanced sampling strategy.
Below we first review general object detection methods, and then long-tail classification methods.

\myparagraph{General object detection}
Deep learning based object detection frameworks are divided into  anchor-based and anchor-free ones.
Anchor-based approaches~\cite{girshick2015region, girshick2015fast, ren2015faster, redmon2018yolov3, lin2017focal} explicitly or implicitly extract features for individual regions thus convert object detection into proposal-level classification which have been largely explored. In contrast, anchor-free approaches focus on detecting key points of objects and construct final detection boxes by properly combining detected key points~\cite{law2018cornernet, duan2019centernet,zhou2019bottom} or expanding the representation of key points~\cite{zhou2019objects,tian2019fcos}. For such detectors, proposal classification is achieved by classifying the key points.

These popular object detection frameworks all employ a softmax classifier for either proposal classification or key-point classification. Our proposed balanced group softmax module can be easily plugged into such mainstream detectors by simply replacing the original softmax classifier.
For simplicity,  we mainly experiment with anchor-based detectors Faster R-CNN~\cite{ren2015faster} and Cascade R-CNN~\cite{cai2018cascade} as well as their corresponding instance segmentation approaches Mask R-CNN~\cite{he2017mask} and HTC~\cite{chen2019hybrid}.

\myparagraph{Long-tail classification}
\label{sec:related_longtail}
Long-tail classification is attracting increasing attention due to its realistic applications.
Current works leverage data re-sampling, cost-sensitive learning, or other techniques.
For data re-sampling methods, training samples are either over-sampled (adding copies of training samples for tail classes)~\cite{han2005borderline},     under-sampled (deleting training samples for head classes)~\cite{drummond2003c4}, or class-balanced sampled~\cite{shen2016relay, mahajan2018exploring}, which motivates RFS~\cite{gupta2019lvis}.
For cost-sensitive learning, the network losses are re-weighted at category level by multiplying different weights on different categories to enlarge the influence of tail-class training samples~\cite{cui2019class, cao2019learning, huang2019deep} or at instance level by multiplying different weights on different training samples for more fine-grained control~\cite{lin2017focal, shu2019meta}.
Some other approaches optimize the classifier trained with long-tail data such as Nearest Class Mean classifier (NCM)~\cite{mensink2013distance, guerriero2018deepncm}, and $\tau$-normalized classifier~\cite{kang2019decoupling}.
These methods are usually sensitive to hyper-parameters and do not perform well when transferred to detection frameworks due to the inherent difference between classification and detection as stated in Sec.~\ref{sec:intro}.

Therefore, an approach specifically designed for long-tail object detection is desirable, and our work is the first successful attempt to overcome classifier imbalance through group-wise training without extra sampling from tail classes.

\begin{figure*}
\begin{center}
\vspace*{-20pt}
	\includegraphics[width=\linewidth]{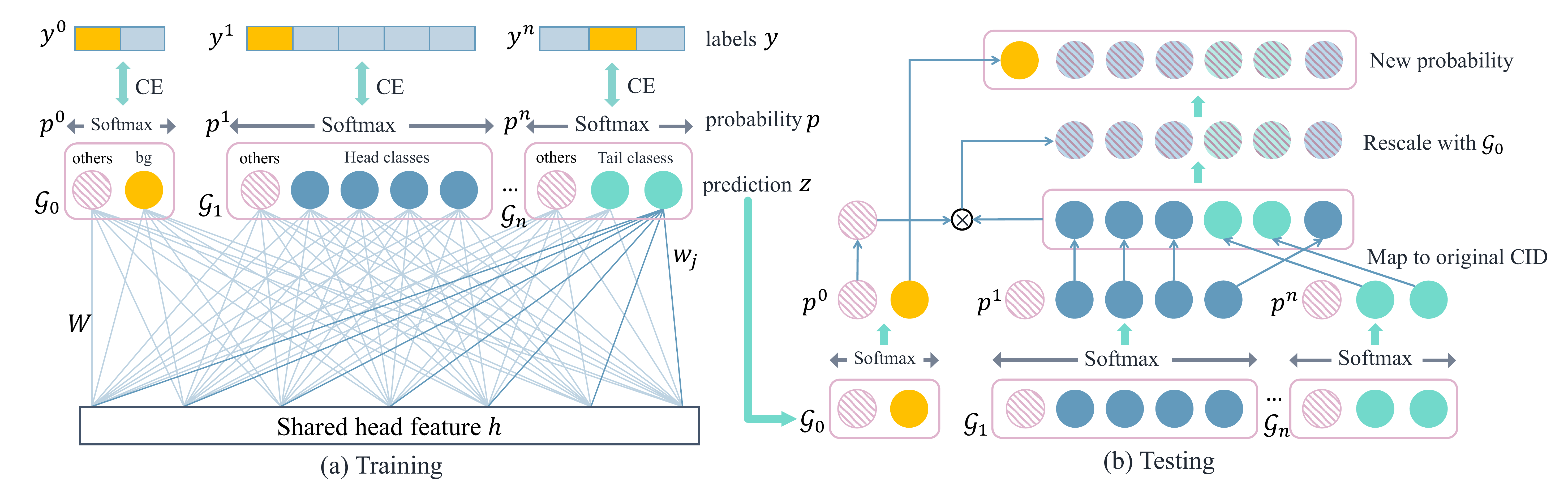}
\vspace*{-15pt}
	\caption{Framework of our balanced group softmax module. (a) Training: Classes containing similar training instances are grouped together. Class \emph{others} is added to each group. $\mathcal{G}_0$ denotes \emph{background} group. Softmax cross entropy (CE) loss is applied to each group individually. (b) Testing: With new prediction $z$, softmax is applied in each group, and probabilities are ordered by their original category id (CID) and re-scaled with foreground probability, generating new probability vectors for following post process. }
	\label{fig:framework}
\vspace*{-20pt}
\end{center}
\end{figure*}

\vspace{-5pt}
\section{Preliminary and Analysis}

\subsection{Preliminary}
\label{sec:preliminary}
We first revisit the popular two-stage object detection framework, by taking   Faster R-CNN~\cite{ren2015faster} as an example. We adopt such a two-stage framework to develop and implement our idea. 

The backbone network $f_\mathrm{back}$ takes an image $I$ as input, and generates a feature map $F=f_{\mathrm{back}}(I)$.
The feature map is then passed to  ROI-align~\cite{he2017mask} or ROI-pooling~\cite{girshick2015fast} to produce $K$ proposals with their own feature $F_k=\mathrm{ROIAlign}(F, b_k)$. Here  $b_k$ denotes proposal $k$.
The classification head $f_{\mathrm{head}}$  then extracts a $d$-dimensional feature $h=f_\mathrm{head}(F_k)$ for each of the proposals.
 Finally, one FC (fully connected) layer is used to transfer $h$ to the ($C+1$)-category prediction ($C$ object classes plus \emph{background}) by $z=Wh + b$, where  $W \in \mathbb{R}^{d\times (C+1)}$ is the classifier weights with each column  $w_j \in \mathbb{R}^d$ related to one specific  category $j$, and $b$ is the bias term.

During training, with ground truth label $y \in \{0,1\}^{C+1}$, softmax cross entropy is applied to compute loss for a specific proposal:
\vspace{-10pt}
\begin{align}
\begin{split}
	\mathcal{L}_k(p, y) = -\sum_{j=0}^{C} y_j \log \left(p_{j}\right),
\end{split}
\end{align}
\vspace{-2em}
\begin{align}
\begin{split}
	p_{j}=\operatorname{softmax}\left(z_{j}\right)=\frac{e^{z_{j}}}{\sum_{i=0}^{C} e^{z_{i}}}.
\end{split}
\label{eq:softmax}
\end{align}
Here $z_{j}$ denotes the $i$-th element of $z$ and 
$p_j$ is the predicted probability of  the proposal being an instance of category $j$.

\subsection{Analysis}
Current well-performing detection models often fail to recognize tail classes when the training set follows a long-tailed distribution. In this section, we try to investigate the underlying mechanism behind such performance drop from balanced dataset to long-tailed dataset, by conducting contrast experiments on their representative examples, \textit{i.e.}, COCO and LVIS.

We adopt a Faster R-CNN~\cite{girshick2015fast} model with R50-FPN backbone. By directly comparing the mAP on the two datasets, the performance drops notably from 36.4\%(COCO) to 20.9\%(LVIS). 
Despite the unfairness as LVIS contains much more classes than COCO (1230 v.s. 80), we can still draw some interesting observations. On head classes, the LVIS model achieves comparable results with COCO. However, when it comes to tail classes, the performance decreases to 0 rapidly. Such a phenomenon implies current detection models are indeed challenged by data imbalance.
To further investigate how the performance degradation is induced by data imbalance, we decouple the detection framework into proposal feature extraction stage and proposal classification stage, following ~\cite{kang2019decoupling}.

Specifically, following the notations in Sec.~\ref{sec:preliminary}, we deem the operations used to generate $h$ as proposal feature extraction, and the last FC layer and softmax in Eqn.~\eqref{eq:softmax} as a softmax classifier. 
Then, we investigate the correlation between the number of training instances and  the weight norm $\Vert w_j \Vert$ in the classifier for each category. 
The results are visualized in Figure~\ref{fig:motivation}. 
We can see for COCO dataset, most categories contain $10^3-10^4$ training instances (at least $10^2$); and classifier weight norms are also relatively balanced  (0.75-1.25) for all foreground categories~\footnote{Note that the first class is \emph{background}(CID=0).}. 
In contrast, for the LVIS dataset, a weight norm $\Vert w_j \Vert$ is highly related to the number of training instances in the corresponding category $j$; the more training examples there are, the larger weight magnitude it will be.
For the extreme few-shot categories (tail classes), their corresponding weight norms are extremely small, even close to zero. Based on such observations, one  can foresee  that prediction scores for tail classes will be congenitally lower than head classes, and proposals of tail classes will be less likely to be selected  after competing with those of head categories within the softmax computation.
This explains why current detection models often fails on tail classes.

Why would the classifier weights be correlated to the number of training instances per-class?
To answer this question, let us further inspect the training procedure of Faster R-CNN. When proposals from a head class $j$ are selected as training samples, $z_j$ should be activated, while the predictions for other categories should be suppressed.
As the training instances for head classes are much more than those of tail classes (\eg, 10,000 vs.\ 1 in some extreme cases), classifier weights of tail classes are much more likely (frequent) to be suppressed by head class ones, resulting in imbalanced weight norms after training.

Therefore, one may see why re-sampling method~\cite{gupta2019lvis, wang2019classification} is able to benefit tail classes on long-tail instance classification and segmentation.
It simply increases the sampling frequency of tail class proposals during training so that the weights of different classes can be equally activated or suppressed, thus balance the tail and head classes to some degree. 
Also,  loss re-weighting methods~\cite{cui2019class, cao2019learning, huang2019deep, lin2017focal, shu2019meta} can take effect in a similar way. 
Though the resampling strategy is able to alleviate data imbalance, it actually introduces new risks like overfitting to tail classes and extra computation overhead. Meanwhile, loss re-weighting is sensitive to per-class loss weight design, which usually varies across different frameworks, backbones and datasets, making it hardly deployable in real-world applications. 
Moreover, re-weighting based methods cannot handle the \emph{background} class well in detection problems. Therefore, we propose a simple yet effective solution  to balance the classifier weight norms without heavy hyper-parameter engineering.

\section{Balanced Group Softmax}
Our novel balanced group softmax module is illustrated in Figure~\ref{fig:framework}. We first elaborate on its formulation and then explain the design details. 

\subsection{Group softmax}
As aforementioned, detector performance is harmed by the positive correlation between weight norms and number of training examples. 
To solve this problem, we propose to divide classes into several disjoint groups and perform the softmax operation separately, such that only classes with    similar numbers of training instances are competing with each other within each group. 
In this way, classes containing significantly different numbers of instances can be isolated from each other during training. The classifier weights of tail classes would not be substantially suppressed by head classes.

Concretely, we divide all the $C$ categories into $N$ groups according to their training instance numbers.
We assign category $j$ to group $\mathcal{G}_n$ if
\vspace{-9pt}
\begin{align}
\begin{split}
s_n^l \leq \mathcal{N}(j) < s_n^h, ~~~~ n > 0
\end{split}
\vspace{-16pt}
\end{align}
where $\mathcal{N}(j)$ is the number of ground-truth bounding boxes for category $j$ in the training set, and $s_n^l$ and $s_n^h$ are hyper-parameters that determine minimal and maximal instance numbers for  group $n$. 
In this work, we set $s_{n+1}^l = s_n^h$ to ensure there is no overlap between groups, and each category can only be assigned to one group.
$N$ and $s_n^l$ are set empirically to make sure that categories in each group contain similar total numbers of training instances.
Throughout this paper, we set $N=4, s_1^l=0, s_2^l=10, s_3^l=10^2, s_4^l=10^3, s_4^h=+\infty$.

Besides, we manually set the $\mathcal{G}_0$ to contain only the \emph{background} category, because it owns the most training instances (typically 10-100 times more than object categories).
We adopt sigmoid cross entropy loss for $\mathcal{G}_0$ here because it only contains one prediction, while for the other groups we use softmax cross entropy loss.
The reason for choosing softmax is that the softmax function inherently owns the ability to suppress each class from another, and less likely produce large numbers of false positives.
During training, for a proposal $b_k$ with ground-truth label $c$, two groups will be activated, which are background group $\mathcal{G}_0$ and foreground group $\mathcal{G}_n$ where $c \in \mathcal{G}_n$.

\subsection{Calibration via category ``others''}

However, we find the above group softmax design suffers from the following issue. During testing, for a proposal, all groups will be used to predict since its category is unknown.
Thus, at least one category per group will receive a high prediction score, and it will be hard  to decide which group-wise prediction we should take, leading to a large number of false positives.
To address this issue, we add a category \emph{others} into every group to calibrate predictions among groups and suppress false positives.
This category \emph{others} contains categories not included in the current group, which can be either \emph{background} or foreground categories in other groups.
For $\mathcal{G}_0$, category \emph{others} also represents foreground classes.
To be specific, for a proposal $b_k$ with ground-truth label $c$, the new prediction $z$ should be $z \in \mathbb{R}^{(C+1)+(N+1)}$.
The probability of class $j$ is calculated by 
\vspace{-5pt}
\begin{align}
\begin{split}
p_j = \frac{e^{z_{j}}}{\sum_{i \in \mathcal{G}_n} e^{z_{i}}}, ~\{n~|~j \in \mathcal{G}_n\}.
\end{split}
\label{eq:newp}
\end{align}
The ground-truth labels should be re-mapped in each group.
In groups where $c$ is not included, class \emph{others} will be defined as the ground-truth class. Then the final loss function is
\vspace{-1.5em}
\begin{align}
\begin{split}
	\mathcal{L}_k= -\sum_{n=0}^N \sum_{i \in \mathcal{G}_n} y_i^n \log \left(p_{i}^n\right),
\end{split}
\end{align}
where $y^n$ and $p^n$ represent the label and probability in $\mathcal{G}_n$.

\subsection{Balancing training samples in groups}
In the above treatment, the newly added category \emph{others} will again become a dominating outlier with overwhelming many instances.
To balance training sample number per group, we only sample a certain number of \emph{others} proposals for training,  which is controlled by a sampling ratio $\beta$.
For $\mathcal{G}_0$, all training samples of \emph{others} will be used since the number of \emph{background} proposals is very large.
For $\{\mathcal{G}_n~|~n\in \mathbb{R}, 1\leq n \leq N\}$, $m_n$ \emph{others} instances will be randomly sampled from all \emph{others} instances, where $m_n=\beta \sum_{i \in \mathcal{G}_n}\mathcal{N}_{batch}(i)$.  $\beta \in [0, +\infty)$ is a hyper-parameter and we conduct an ablation study in Sec.~\ref{sec:ablation_beta} to show the impact of $\beta$. Normally, we set $\beta=8$.
$\mathcal{N}_{batch}(i)$ indicates the instances for category $i$ in current batch.

Namely, within the groups that contain the ground-truth categories, \emph{others} instances will be sampled proportionally based on a mini-batch of $K$ proposals.
If there is no normal categories activated in one group, all the \emph{others} instances will not be activated. This group is ignored.
In this way, each group can keep balanced with a low ratio of false positives. 
Adding category \emph{others} brings 2.7\% improvement over the baseline.

\subsection{Inference}
During inference, we first generate $z$ with the trained model, and apply softmax in each group using Eqn.~ (\ref{eq:newp}).
Except for $\mathcal{G}_0$, all nodes of \emph{others} are ignored, and probabilities of all categories are ordered by the original category IDs.
$p_0^0$ in $\mathcal{G}_0$ can be regarded as the probability of foreground proposals.
Finally, we rescale all probabilities of normal categories with $\widetilde{p_j} = p_0^0 \times p_j$.
This new probability vector is fed to the following post-processing steps like NMS
to produce final detection results.
It should be noticed that the $\widetilde{p}$ is not a real probability vector technically since the summation of it does not equal to 1.
It plays the role of the original probability vector which guides the model through selecting final boxes.

\section{Experiments}

\subsection{Dataset and setup}
We conduct  experiments on the recent Large Vocabulary Instance Segmentation (LVIS) dataset~\cite{gupta2019lvis}, which contains 1,230 categories  with both bounding box and instance mask annotations.
For object detection experiments, we use only bounding box annotation for training and evaluation. 
When exploring BAGS's generalization to instance segmentation, we use mask annotations.
Please refer to our supplementary materials for implementation details.

Following~\cite{wang2019classification}, we split the categories in the  validation set of LVIS into 4 bins according to their  training instance numbers  to  evaluate the model performance on the head and tail classes more clearly.
Bin$_i$ contains  categories that have $10^{i-1}$ to $10^i$ instances.
We refer categories in the first two bins as ``tail classes", and categories in the other two bins as ``head classes".
Besides the official metrics mAP, AP$_r$ (AP for rare classes), AP$_c$ (AP for common classes), and AP$_f$ (AP for frequent classes) that are provided by LVIS-api\footnote{https://github.com/lvis-dataset/lvis-api}, we also report AP on different bins.
AP$_i$ denotes the mean AP over the categories from Bin$_i$.

\begin{table*}[]
\small
\renewcommand{\tabcolsep}{2.0pt}
\renewcommand{\arraystretch}{1.3}
\vspace*{-20pt}
\begin{center}
\begin{tabular}{cl|c|cccc|ccc|c|ccccc}
\toprule
ID & Models              & \emph{\textbf{mAP}}     & AP$_1$  & AP$_2$  & AP$_3$  & AP$_4$  & AP$_r$  & AP$_c$  & AP$_f$  & ACC     & ACC$_1$ & ACC$_2$ & ACC$_3$ & ACC$_4$ & ACC$_{bg}$ \\
\midrule
(1) & R50-FPN              & \emph{20.98} & 0.00  & 17.34 & 24.00 & 29.99 & 4.13  & 19.70 & 29.30 & 92.78 & 0.00  & 2.47  & 25.30 & 45.87 & 95.91    \\
(2) & x2               & \emph{21.93} & 0.64  & 20.94 & 23.54 & 28.92 & 5.79  & 22.02 & 28.26 & 92.62 & 0.00  & 5.60  & 26.51 & 45.71 & 95.69    \\
(3) & Finetune tail            & \emph{22.28} & 0.27  & 22.58 & 23.89 & 27.43 & 5.67  & 23.54 & 27.34 & 94.81 & 0.00  & 5.04  & 5.58  & 5.86  & 99.85    \\
\hline
(4) & RFS~\cite{gupta2019lvis}           & \emph{23.41} & 7.80  & 24.18 & 23.14 & 28.33 & 14.59 & 22.74 & 27.77 & 92.71 & 0.60  & 7.50  & 25.62 & 44.39 & 95.84    \\
(5) & RFS-finetune         & \emph{22.66} & 8.06  & 23.07 & 22.43 & 27.73 & 13.44 & 22.06 & 27.09 & 92.77 & 0.60  & 7.14  & 25.08 & 43.79 & 95.91    \\
\hline
(6) & Re-weight           & \emph{23.48} & 6.34  & 22.91 & 23.88 & \textbf{30.12} & 11.47 & 22.41 & 29.61 & \textbf{94.84} & 0.00  & 0.82  & 9.57  & 17.40 & 99.53    \\
(7) & Re-weight-cls       & \emph{24.66} & 10.04  & 24.12 & 24.57 & 31.07 & 14.16 & 23.51 & \textbf{30.28} & 94.76 & 0.00  & 0.34  & 7.72  & 16.02 & 99.64    \\
(8) & Focal loss~\cite{lin2017focal}          & \emph{11.12} & 0.00  & 10.24 & 13.36 & 13.17 & 2.74  & 11.13 & 14.46 & 3.87  & 0.00  & 17.45 & 40.11 & \textbf{49.31} & 1.35     \\
(9) & Focal loss-cls      & \emph{19.29} & 1.64  & 19.30 & 20.64 & 23.70 & 6.60  & 19.81 & 23.71 & 2.90  & 0.00  & 27.67 & \textbf{48.53} & 48.89 & 0.16     \\

\hline
(10) & NCM-fc~\cite{kang2019decoupling}             & \emph{16.02} & 5.87  & 14.13 & 16.97 & 21.40 & 10.31 & 13.92 & 20.92 & 94.29 & 0.00  & 0.02  & 0.23  & 0.15  & 100.00   \\
(11) & NCM-conv~\cite{kang2019decoupling}             & \emph{12.56} & 4.20  & 9.71  & 13.75 & 18.46 & 6.11  & 10.39 & 17.85 & 94.29 & 0.00  & 0.00  & 0.20  & 0.10  & 100.00   \\
\hline
(12) & $\tau$-norm~\cite{kang2019decoupling}          & \emph{11.01} & 0.00  & 11.71 & 12.01 & 12.36 & 2.07  & 12.30 & 12.97 & 5.91  & 0.00  & \textbf{30.32} & 39.49 & 49.14 & 3.42     \\
(13) & $\tau$-norm-select          & \emph{21.61} & 0.35  & 20.07 & 23.43 & 29.16 & 6.18  & 20.99 & 28.54 & 92.43 & 0.00  & 13.19 & 20.62 & 38.98 & 95.91    \\
\midrule
(14) & \textbf{Ours}                & \emph{\textbf{25.96}} & \textbf{11.33} & \textbf{27.64} & \textbf{25.14} & 29.90 & \textbf{17.65} & \textbf{25.75} & 29.54 & 93.71 & \textbf{2.06}  & 7.50  & 22.07 & 35.88 & 97.41   \\
\bottomrule
\end{tabular}
\end{center}
\vspace{-5pt}
\caption{Comparison with state-of-the-art methods transferred from long-tail image classification on LVIS \emph{val} set. \textbf{Bold} numbers denote the best results among all models. Model (1) and (4) are initialized with model pre-trained on COCO dataset. All the others are initialized with model (1). ``-cls" denotes only train the classification FC layer $W$ and $b$, and the other parameters are frozen. Model (10) and (11) represent NCM model using classification FC features and ROI-pooled Conv feature to calculate category centers respectively. Model (13) means using $\tau$-norm results only on foreground proposals. Refer to our supplementary materials for more implementation details.} 
\vspace*{-15pt}
\label{tab:compare_others}
\end{table*}

\subsection{Main results on LVIS}

We transfer multiple state-of-the-art methods for long-tail classification to the Faster R-CNN framework, including fine-tuning on tail classes, repeat factor sampling (RFS)~\cite{mahajan2018exploring}, category loss re-weighting, Focal Loss~\cite{lin2017focal}, NCM~\cite{kang2019decoupling,snell2017prototypical}, and $\tau$-normalization~\cite{kang2019decoupling}. We carefully adjust the hyperparameter settings to make them suitable for object detection. Implementation details are provided in our supplementary material.  We report their detection performance and  proposal classification accuracy in Table \ref{tab:compare_others}.

\emph{How well does naive baseline perform?} We take Faster R-CNN with ResNet-50-FPN backbone as the baseline (model (1) in the table) that achieves 20.98\% mAP but   0  AP$_1$. 
The baseline model misses   most tail categories due to domination of other classes.
Consider other  models are   initialized by model (1) and further fine-tuned by another 12 epochs. To make sure the improvement is not from longer training schedule, we train model (1) with   another 12 epochs for fair comparison. This gives model (2). Comparing model (2) with model (1), we find longer training   mainly improves on AP$_2$, but AP$_1$ remains around 0. Namely, longer training hardly helps improve the performance for low-shot categories with instances less than 10.
Fine-tuning model (1) on tail-class training samples (model (3)) only increases  AP$_2$   notably while decreases AP$_4$   by 2.5\%, and AP$_1$ remains  0.
This indicates the original softmax classifier cannot perform well when the number of training instances is too small.

\emph{Do long-tail classification methods help?}  
We observe sampling-based method RFS (model (4)) improves the overall mAP by 2.5\%. 
The AP for tail classes is improved while maintaining AP for head classes.
However, RFS 
increases the training  time cost by $1.7\times$.
We also try to initialize the model with model (1), obtaining model (5).
But mAP drops by 0.8\% due to   over-fitting.

For cost sensitive learning methods, 
both model (6) and (7) improve the performance, while model (7) works better.
This confirms the observation in~\cite{kang2019decoupling}  that decoupling feature learning and classifier benefits long-tail recognition still applies for object detection.
For focal loss, we directly apply sigmoid focal loss  at proposal level.
It is worth noting that in terms  of proposal classification, the accuracy over all the object classes (ACC$_{1,2,3,4}$) increases notably.
However, for \emph{background} proposals, ACC$_{bg}$ drops from 95.8\% to 0.16\%, leading to a large number of false positives and low AP.
This phenomenon again highlights the difference between long-tail detection and classification\textemdash
the very special \emph{background} class should be carefully treated.

For NCM, we try to use both FC feature just before classier (model (10)), and Conv feature extracted by ROI-align (model (11)).
However, our observation is NCM works well for extremely low-shot classes, but is not good for head classes.
Moreover, NCM can provide a good 1-nearest-neighbour classification label. But for detection, we also need the whole probability vector to be meaningful so that
scores of different proposals on the same categories can be used to evaluate the quality of proposals.

The $\tau$-normalization model (12) suffers  similar challenge as Focal Loss model (8).
The many-shot \emph{background} class is extremely dominating.
Though foreground proposal accuracy is greatly increased, ACC$_{bg}$ drops hugely.
Consequently, for model (13), the 
  proposals     categorized to \emph{background}   inherit   prediction of the original model while the others take $\tau$-norm results.
However, the improvement is limited.
We should notice that AP$_1$ and ACC$_1$ are still 0 after $\tau$-norm, but AP$_2$ and ACC$_2$ are improved. 

\emph{How well does our method perform?} For our model,
except for $\mathcal{G}_0$, we split the normal categories into 4 groups for group softmax computation, with $s^l$ and $s^h$ to be (0, 10), (10, $10^2$), ($10^2$, $10^3$), ($10^3$, $+\infty$) respectively, and 
$\beta=8$.
Our model is initialized with model (1) 
, and the classification FC layer is randomly initialized since the output shape changed.
Only this FC layer is trained for another 12 epochs, and all other parameters are frozen.
Our results surpass all the other methods by a large margin.
AP$_1$ increases 11.3\%, AP$_2$ increases 10.3\%, with AP$_3$ and  AP$_4$ almost unchanged.
This result verifies the effectiveness of our designed balanced group softmax module.

\begin{table}[]
\footnotesize
\renewcommand{\tabcolsep}{2.0pt}
\renewcommand{\arraystretch}{1.3}
\vspace*{-15pt}
\begin{center}
\begin{tabular}{c|c|cccc|ccc}
\toprule
Models                & \textbf{\emph{mAP}}     & AP$_1$  & AP$_2$  & AP$_3$  & AP$_4$  & AP$_r$  & AP$_c$  & AP$_f$  \\
\midrule
Faster R50  & \emph{20.98} & 0.00  & 17.34     & 24.00       & 29.99          & 4.13  & 19.70 & 29.30 \\
Ours         & \textbf{\emph{25.96}} & 11.33 & 27.64     & 25.14       & 29.90          & 17.65 & 25.75 & 29.54 \\
\midrule
Faster X101 & \emph{24.63} & 0.79  & 22.37     & 27.45       & 32.73          & 5.80  & 24.54 & 32.25 \\
Ours         & \textbf{\emph{27.83}} & 14.99 & 28.07     & 27.93       & 32.02          & 18.78 & 27.32 & 32.07 \\
\midrule
Cascade X101 & \emph{27.16} & 0.00  & 24.06     & 31.09       & 36.17          & 4.84  & 27.22 & 36.00 \\
Ours         & \textbf{\emph{32.77}} & 19.03 & 36.10     & 31.13       & 34.96          & 28.24 & 32.11 & 35.41 \\
\bottomrule
\end{tabular}
\end{center}
\vspace*{-5pt}
\caption{Results with stronger backbone ResNeXt-101-64x4d and stronger framework Cascade R-CNN. All \emph{Ours} models are initialized with their plain counterparts.}
\vspace*{-15pt}
\label{tab:cascade101}
\end{table}

\emph{Extension of our method to stronger models.} To further verify the generalization of our method, we   change Faster R-CNN backbone to ResNeXt-101-64x4d~\cite{xie2017aggregated}.
Results are shown in Table \ref{tab:cascade101}. 
On this much stronger backbone, our method still gains 3.2\% improvement.
Then, we apply our method to state-of-the-art Cascade R-CNN~\cite{cai2018cascade} framework with changing all 3 softmax classifiers in 3 stages to our BAGS module.
Overall mAP is increased significantly by  5.6\%.
Our method brings persistent gain with 3 heads.

\subsection{Results for instance segmentation}
\label{sec:extend_to_seg}
\begin{table*}[]
\footnotesize
\renewcommand{\tabcolsep}{1.5pt}
\renewcommand{\arraystretch}{1.2}
\vspace*{-15pt}
\begin{center}
\begin{tabular}{ccc|c|cccc|ccc|c|cccc|ccc}
\toprule
ID & Models & Backbone           & \textbf{\emph{mAP}}     & AP$_1$  & AP$_2$  & AP$_3$  & AP$_4$  & AP$_r$  & AP$_c$  & AP$_f$  & \textbf{\emph{mAP$^m$}} & AP$_1^m$ & AP$_2^m$ & AP$_3^m$ & AP$_4^m$ & AP$_r^m$ & AP$_c^m$ & AP$_f^m$ \\
\midrule
(1) & Mask-RFS*~\cite{gupta2019lvis} & R50 	& -- & --  & -- & -- & -- & --  & -- & -- & \emph{24.40} & --   & --  & --  & --  & 14.50   & 24.30  & 28.40  \\
(2) & Mask-RFS*~\cite{gupta2019lvis} & R101 	& -- & --  & -- & -- & -- & --  & -- & -- & \emph{26.00} & --   & --  & --  & --  & 15.80   & 26.10  & 29.80  \\
(3) & Mask-RFS*~\cite{gupta2019lvis} & X101-32x8d	& -- & --  & -- & -- & -- & --  & -- & -- & \emph{27.10} & --   & --  & --  & --  & 15.60   & 27.50  & 31.40  \\
(4) & Mask-Calib*~\cite{wang2019classification} & R50& -- & --  & -- & -- & -- & --  & -- & -- & \emph{21.10} & 8.60   & 22.00 & 19.60  & 26.60 & --  & --  & --   \\
(5) & HTC-Calib*~\cite{wang2019classification} & X101 & -- & --  & -- & -- & -- & --  & -- & -- & \emph{29.85}  & 16.05   & 30.60  & 29.80 & 33.50 & --  & --  & --   \\
(6) & HTC-Calib*~\cite{wang2019classification} & X101-MS-DCN & -- & --  & -- & -- & -- & --  & -- & -- & \emph{32.10} & 12.70   & 32.10  & 33.60 & 37.00  & --  & --  & --   \\
\midrule
(7) & Mask R-CNN & R50        & \emph{20.78} & 0.00  & 15.88 & 24.61 & 30.51 & 3.28  & 18.99 & 30.00 & \emph{20.68} & 0.00   & 17.06  & 23.66  & 29.62  & 3.73   & 19.95  & 28.37  \\
(8) & Ours & R50   & \emph{25.76} & 9.65  & 26.20 & 26.09 & 30.45 & 15.03 & 25.45 & 30.42 & \emph{26.25} & 12.81  & 28.28  & 25.15  & 29.61  & 17.97  & 26.91  & 28.74  \\
\midrule
(9) & HTC &   X101      & \emph{31.28} & 5.02  & 31.71 & 33.24 & 37.21 & 12.39 & 32.58 & 37.18 & \emph{29.28} & 5.11   & 30.34  & 30.62  & 34.37  & 12.11  & 31.32  & 33.58  \\
(10) & Ours &  X101  & \emph{33.68} & 19.95 & 36.14 & 32.82 & 36.06 & 25.43 & 34.12 & 36.42 & \emph{31.20} & 17.33  & 33.87  & 30.34  & 33.29  & 23.40  & 32.34  & 32.89  \\
\midrule
(11) & HTC  & X101-MS-DCN   & \emph{34.61} & 5.80  & 35.36 & 36.87 & 40.50 & 14.24 & 35.98 & \textbf{41.03} & \emph{31.94} & 5.56   & 33.07  & \textbf{33.75}  & \textbf{37.02}  & 13.67  & 34.04  & \textbf{36.62}  \\
(12) & \textbf{Ours} & X101-MS-DCN & \textbf{\emph{37.71}} & \textbf{24.40} & \textbf{40.30} & \textbf{36.67} & \textbf{40.00} & \textbf{29.43} & \textbf{37.78} & 40.92 & \textbf{\emph{34.39}} & \textbf{21.07}  & \textbf{36.69}  & 33.71  & 36.61  & \textbf{26.79}  & \textbf{35.04}  & 36.61  \\
\bottomrule
\end{tabular}
\end{center}
 \vspace*{-5pt}
\caption{Results of bounding box and mask AP when extending our method to instance segmentation on LVIS \emph{val} set. AP$^m$ denotes AP of instance segmentation mask. All backbones are with FPN. X101 denotes X101-64x4d. * results are from the corresponding cited paper. \textbf{Bold} numbers indicate the best results among all models. Model (8)(10)(12) are initialized with model (7)(9)(11).}
\label{tab:segmentaion}
\end{table*}

We further evaluate our method benefits for instance segmentation models including Mask R-CNN~\cite{he2017mask} and state-of-the-art HTC~\cite{chen2019hybrid} on LVIS. Here HTC models are trained with COCO stuff annotations for a segmentation branch.
Results are shown in Table \ref{tab:segmentaion}.
First, comparing our models (8)(10)(12) with their corresponding baseline models (7)(9)(11), mAPs of both bounding box and mask increase largely.
Our models fit the tail classes much better while APs on head classes  drop slightly.
Second, we compare our results with state-of-the-art results~\cite{wang2019classification, gupta2019lvis} on LVIS instance segmentation task.
With both Mask R-CNN framework and ResNet-50-FPN backbone, our model (8) surpass RFS (1) and Calib (4) by at least 1.8\%.
With both HTC framework and ResNeXt-101-FPN backbone, our model (10) is 1.4\% better than Calib (5).
With ResNeXt-101-FPN-DCN backbone and multiscale training, our model (12) is 2.3\% better than Calib (6). Our method establishes new state-of-the-arts in terms of both bounding box and mask criterion.

\begin{figure}
\begin{center}
\vspace*{-15pt}
	\includegraphics[width=\linewidth]{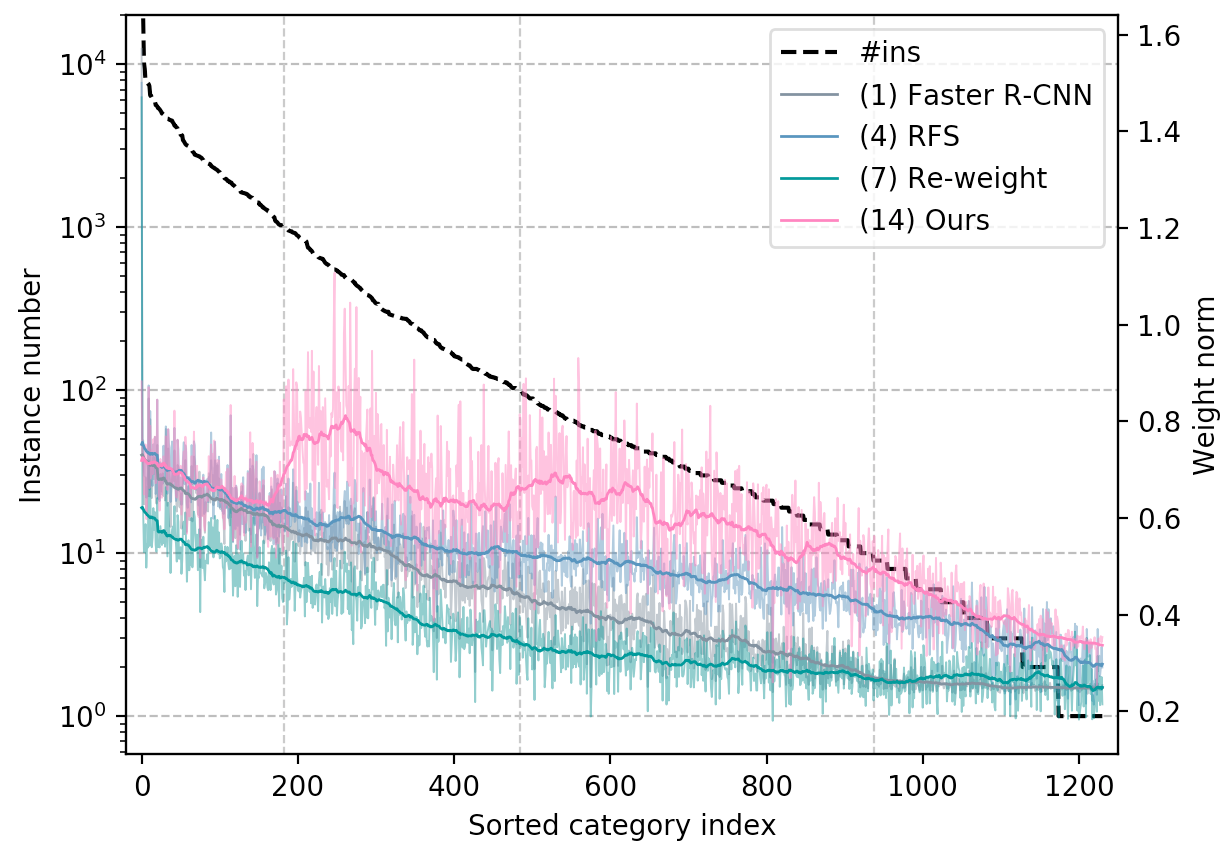}
    \vspace*{-15pt}
	\caption{Comparison of weight norms from model (1)(4)(7)(14) in Table~\ref{tab:compare_others}. The vertical dashed lines split all categories into Bin$_{1,2,3,4}$.}
 	\vspace*{-25pt}
	\label{fig:draw_parames}
\end{center}
\end{figure}

\subsection{Model analysis}
\emph{Does our   method balance classifiers well?}
We visualize the classifier weight norm $W$ of model (1)(4)(7) and our model (14) of Table~\ref{tab:compare_others} in Figure~\ref{fig:draw_parames}.
Weights of RFS on tail classes are obviously enlarged.
Re-weighting method suppresses the weights of head classes and lifts weights of the tail classes.
For ours, since we decouple the relationships of different group of categories, weights of $\mathcal{G}_1, \mathcal{G}_2$ and $\mathcal{G}_3$ are almost at the same level.
Though weights of $\mathcal{G}_4$ are still smaller, they have been better balanced than the original model.
Noting that the weights norm of our model are less related to the training instance numbers in each group, implying such decoupling  benefits network training.

\emph{How much \emph{background} and \emph{others} contribute?}
See Table~\ref{tab:binnum}. With baseline model (0), 
directly grouping normal categories to 4 sets without adding \emph{background} $\mathcal{G}_0$   and \emph{others}  in each group, we get results of (1).
For model (1), during inference, scores of each group are fed to softmax respectively, and concatenated directly for NMS.
Though AP$_1$ improves 5.7\%, performance on all the other Bins drops significantly.
This is because we do not have any constraints for FPs. For  a single  proposal, at least one category will be activated in each group, leading to many FPs.
When we add $\mathcal{G}_0$ (model (2)), and use $p_0^0$ to rescale scores of normal categories, we get 1.9\% improvement over model (1), but still worse than model (0).
For model (3), we add category \emph{others} into each groups, and not using $\mathcal{G}_0$, we obtain 2.7\% performance gain.

\begin{table}[]
\small
\vspace*{-10pt}
\renewcommand{\tabcolsep}{1.5pt}
\renewcommand{\arraystretch}{1.2}
\begin{center}
\begin{tabular}{cccc|c|cccc|ccc}
\toprule
ID       & b                       & o                    & N & \textbf{\emph{mAP}}     & AP$_1$  & AP$_2$  & AP$_3$  & AP$_4$  & AP$_r$  & AP$_c$  & AP$_f$  \\
\midrule
(0) &                           &                           &         & \emph{20.98} & 0.00  & 17.34 & 24.00 & 29.99 & 4.13  & 19.70 & 29.30 \\
\hline
(1)          &                           &                           & 4       & \emph{17.82} & 5.71  & 17.07 & 18.09 & 23.13 & 8.52  & 17.44 & 22.01 \\
(2)          & \checkmark &                           & 4       & \emph{19.73} & 7.18  & 19.66 & 18.80 & 25.95 & 9.89  & 19.32 & 24.19 \\
(3)          &                           & \checkmark & 4       & \emph{23.74} & 9.90  & 24.06 & 23.38 & 28.88 & 15.46 & 22.58 & 28.49 \\
(4)          & \checkmark & \checkmark & 2       & \emph{25.31} & 6.53  & 27.55 & 24.19 & 30.35 & 15.30 & 25.14 & 29.53 \\
(5)          & \checkmark & \checkmark & 4       & \emph{25.96} & 11.33 & 27.64 & 25.14 & 29.90 & 17.65 & 25.75 & 29.54 \\
(6)          & \checkmark & \checkmark & 8       & \emph{24.85} & 7.79 & 26.05 & 24.59 & 29.58 & 14.11 & 24.79 & 29.21 \\
\bottomrule
\end{tabular}
\end{center}
 \vspace*{-5pt}
\caption{Effect of adding different components to our module. \emph{b} for \emph{background} group $\mathcal{G}_0$. \emph{o} for adding category \emph{others} to all bins. \emph{N} is number of normal category groups.}
\vspace*{-16pt}
\label{tab:binnum}
\end{table}

\emph{How many groups to use in BAGS?}
With rescaling with $\mathcal{G}_0$, another 2.2\% improvement is obtained (model (5)).
If we reduce the group number from 4 to 2, as shown in model (4), the overall mAP drops 0.6.
However, specifically, it should be noticed that AP$_1$ becomes much worse, while AP$_4$ increases a little.
Using more groups does not help as well (model(6)).
Since \#ins for Bin$_1$ is too small for $N=4$,  dividing Bin$_1$ into 2 bins further decreases \#ins of per group, leading to highly insufficient training for tails.
To sum up, adding category \emph{others} into each group matters a lot, and using specially trained $p_0^0$ to suppress \emph{background} proposals works better than just \emph{others}.
Finally, grouping categories into bins and decoupling the relationship between tail and head classes benefits a lot for learning of tail classes.

\label{sec:ablation_beta}

\begin{figure}[!htb]
\vspace*{-15pt}
\begin{center}
	\includegraphics[width=0.8\linewidth]{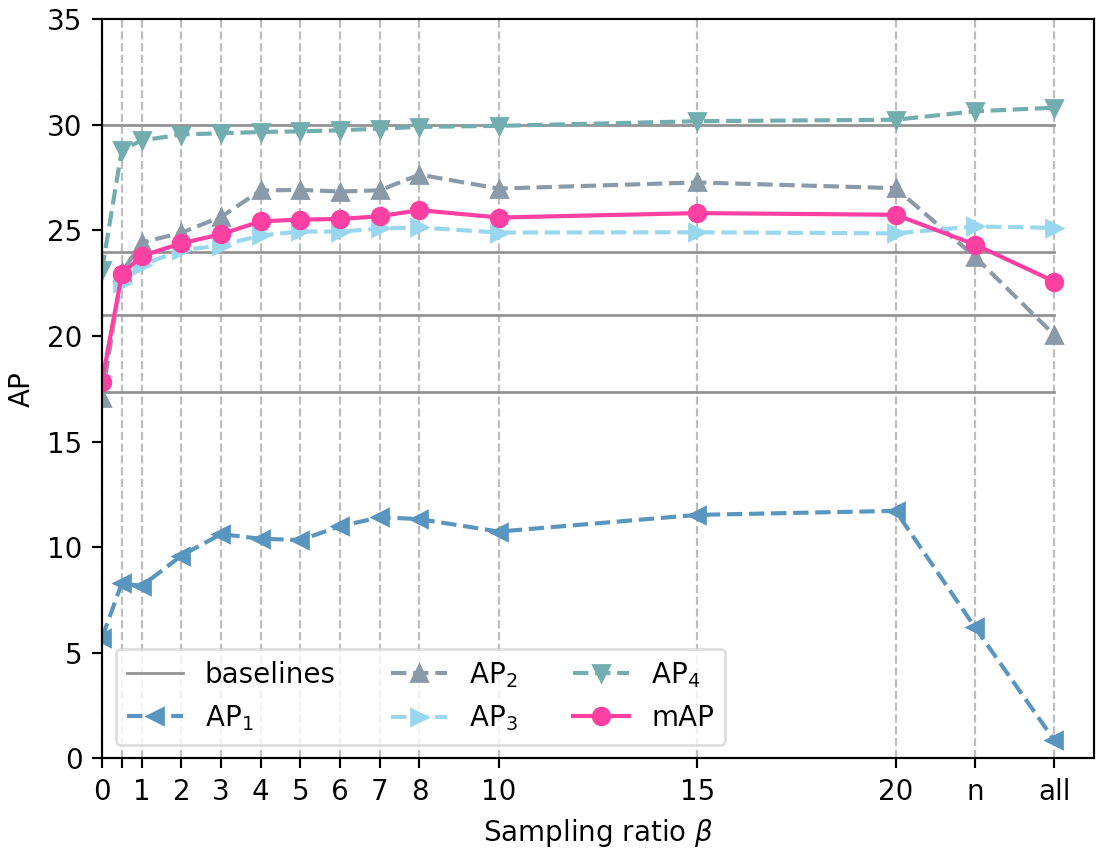}
 	\vspace*{-5pt}
	\caption{Influence of sample ratio $\beta$. \emph{n} indicates all \emph{others} in the activated groups, and \emph{all} indicates all \emph{others} in all the groups.}
 	\vspace*{-20pt}
	\label{fig:bgn}
\end{center}
\end{figure}

\emph{Impact of $\beta$ in BAGS.} After adding category \emph{others} to all groups, we need to sample training instances for \emph{others}.
Using all \emph{others} proposals will lead to imbalance problem in each group.
Thus, our strategy is to sample \emph{others} with ratio $\beta$, so that \#ins \emph{others}$:$\#ins \emph{normal} = $\beta$.
As shown in Fig.\ref{fig:bgn}, mAP continuously increases when we increase $\beta$ until $\beta=8$.
If we use all \emph{others} proposal in activated group (indicated as $n$ in x-axis), the performance for head classes keep increasing, but that for tail classes drops a lot.
If we train all \emph{others} proposals no matter whether there are normal categories being activated (indicated as $all$ in x-axis), mAP gets worse.
This verifies our opinion that another imbalance problem could worsen the results.


\subsection{Results on COCO-LT}
To further verify the generalization ability of our method, we construct a long-tail distribution COCO-LT dataset by sampling images and annotations from COCO~\cite{lin2014microsoft}.
We get similar results on COCO-LT as on LVIS.
Our model still introduces over 2\% improvement on mAP (+2.2\% for Faster R-CNN, +2.4\% for Mask R-CNN bounding box, +2.3\% for Mask R-CNN mask), especially gaining large improvement on tail classes (from 0.1\% to 13.0\% for bounding box) with both Faster R-CNN and Mask R-CNN frameworks.
Please refer to our supplementary materials for dataset construction, data details, and full results.

\section{Conclusion}
\vspace{-5pt}
In this work, we first reveal a reason for poor detection performance   on long-tail data is that the classifier  becomes imbalanced due to insufficiently training on low-shot classes,  by analyzing their classifier weight norms.
Then, we investigate multiple  solid baseline methods transferred from long-tail  classification, but we found they are limited in addressing challenges for the detection task. 
We thus propose a balanced group softmax module to undertake the imbalance problem of classifiers, which achieves notably better results on different strong backbones  for long-tail detection as well as instance segmentation.

\clearpage

\section{Supplementary materials}
\subsection{Implementation details}

\subsubsection{Experiment setup}
Our implementations are based on the MMDetection toolbox~\cite{mmdetection} and Pytorch~\cite{paszke2017automatic}.
All the models are trained with 8 V100 GPUs, with a batch size of 2   per GPU, except for HTC models (1 image per GPU).
We use SGD optimizer with learning rate = 0.01, and decays twice at the $8_{th}$ and $11_{th}$ epochs with factor = 0.1.
Weight decay = 0.0001.
Learning rate warm-up are utilized.
All \emph{Ours} models are initialized with their corresponding baseline models that directly trained on LVIS with softmax, and only the last FC layer is trained of another 12 epochs, with learning rate = 0.01, and decays twice at the $8_{th}$ and $11_{th}$ epochs with factor = 0.1.
All other parameters are frozen.
  
\subsubsection{Transferred methods}
Here, we elaborate on the detailed implementation for transferred long-tail image classification methods in Table 1 of the main text.

\paragraph{Repeat factor sampling (RFS)}
\emph{RFS}~\cite{mahajan2018exploring} is applied to LVIS instance segmentation in~\cite{gupta2019lvis}.
It increases the sampling rate for tail class instances by oversampling images  containing these categories.
We implement RFS with its best practice settings given by~\cite{gupta2019lvis} with $t=0.001$.

\paragraph{Re-weight}
\emph{Re-weight} is a category-level cost sensitive learning method.
Motivated by~\cite{cui2019class}, we re-weight losses of different categories according to their corresponding number of training instances.
We calculate $\{\alpha_j = 1/{\mathcal{N}(j)}\vert j\in [1, 2, ..., C]\}$, where $\mathcal{N}(j)$ denotes the number of instance for category $j$. We normalize $\alpha_j$ by dividing the mean of all $\alpha$, namely $\mu_\alpha$ and cap their values between 0.01 and 5.
$\alpha_0$ is set to 1 for \emph{background} class.
The model (6) and (7) are both initialized with model (1).
Model (6) fine-tunes all parameters in the network except for Conv1 and Conv2.
Model (7) only fine-tunes the last fully connected layer, namely $W$ and $b$ in Sec.3.1 in the main text, and $\beta$ is set to 0.999.

\begin{figure}[t]
\begin{center}
   \includegraphics[width=\linewidth]{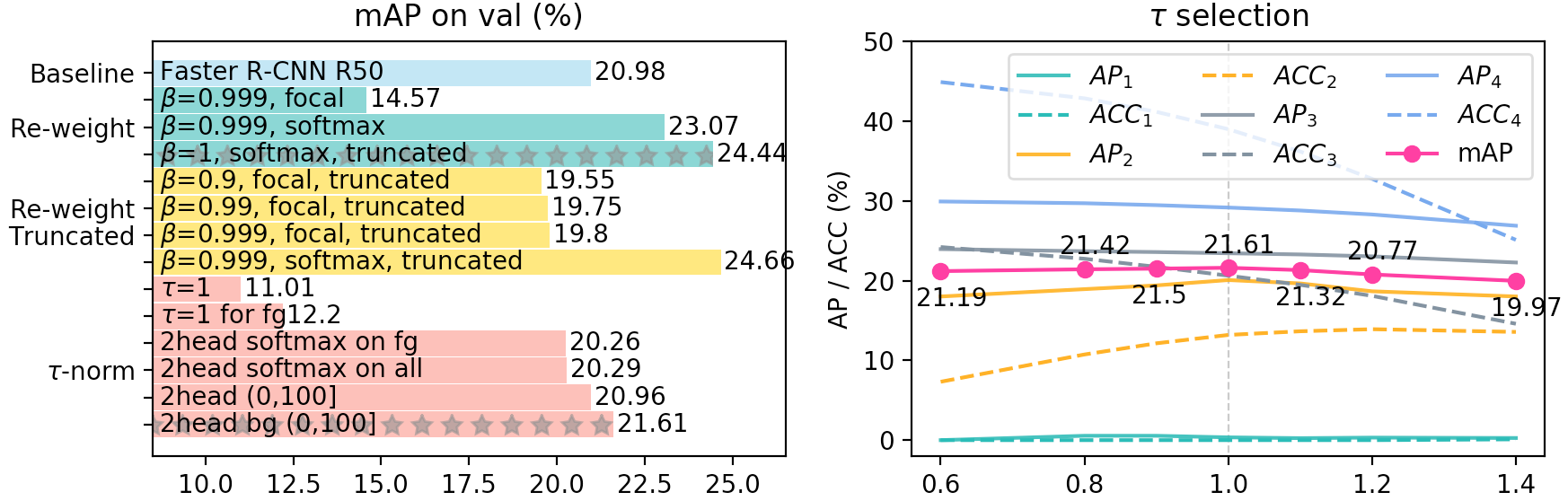}
\end{center}
   \caption{Settings we tried for~\cite{cui2019class} and \cite{kang2019decoupling}.}
\label{fig:results}
\end{figure}

Fig.\ref{fig:results} left shows more settings we have tried for loss re-weighting.
we tried ~\cite{cui2019class}'s best practice \{$\beta$=0.999, \emph{focal}, $\gamma$=0.5\}  by setting \#bg=3$\times$\#fg, but only got 14.57\% mAP.
\{$\beta$=0.999, \emph{softmax}\}=23.07\% indicates softmax works better for Faster R-CNN.
So our (6) in Tab.1 are improved version of \{$\beta$=1,~\emph{softmax}\} with weights truncated to [0.01,5].
We further try to add weight truncation to $\beta$=\{0.9, 0.99, 0.999\}, loss=\{\emph{softmax, ~focal}\}, 
and set $w_{bg}$=1, $\gamma$=2 (loss for $\gamma$=0.5 is too small), and finally found that \{$\beta$=1,~\emph{softmax},~truncated\} (model 7) works best.

\paragraph{Focal loss}
\emph{Focal loss}~\cite{lin2017focal} re-weights the cost at image-level for classification.
We directly apply Sigmoid focal loss  at proposal-level.
Similar to models (6) and (7), models (8) and (9) are initialized with model (1). Then we finetune the whole backbone and classifier ($W, b$) respectively.

\paragraph{Nearest class mean classifier (NCM)}
\emph{NCM} is another commonly used approach that first computes the mean feature for each class on training set.
During inference, 1-NN algorithm is applied with cosine similarity on $L_2$ normalized mean features~\cite{kang2019decoupling,snell2017prototypical}.
Thus, for object detection, with the trained Faster R-CNN model (1), we first calculate the mean feature for proposals of each class on training set except for \emph{background} class.
At inference phase, features for all the proposals are extracted. We then calculate cosine similarity of all the proposal features with the class centers.
We apply softmax over similarities of all categories to get a probability vector $p_n$ for normal classes.
To recognize background proposals we directly take the probability $p_0$ of background class from model (1), and update $p_n$ with $p_n \times (1-p_0)$.
We try both FC feature just before classier (model (10)), and Conv feature  extracted by ROI-align (model (11)) as proposal features.

\paragraph{$\tau$-normalization}
$\tau$-normalization~\cite{kang2019decoupling} directly scale the classifier  weights $W=\{w_j\}$  by $\widetilde{w_{i}}=\frac{w_{i}}{\left\|w_{i}\right\|^{\tau}}$, where $\tau \in (0,1)$ and $\Vert\cdot\Vert$ denotes $L_2$ norm.
It achieves state-of-the-art performance on long-tail classification~\cite{kang2019decoupling}.
For model (13), we first obtain results from both the original model and the $\tau$-normed model.
 The original model is good at categorizing background. Thus, if the proposal is categorized to \emph{background} by the original model, we select the results of the original model for this proposal. Otherwise, the $\tau$-norm results will be selected.
In spite of this, we designed multiple ways to deal with \emph{bg} (background class) (Fig~\ref{fig:results} red bars), and found the above way perform best. 
We also searched $\tau$ value on \emph{val} set, and found $\tau$=1 is the best (Fig~\ref{fig:results} right).

\subsection{How to train our  model}

\begin{table}[]
\footnotesize
\renewcommand{\tabcolsep}{1.8pt}
\renewcommand{\arraystretch}{1.3}
\begin{center}
\begin{tabular}{ccc|c|cccc|ccc}
\toprule
ID & Mode & Part & \textbf{mAP}     & AP$_1$ & AP$_2$  & AP$_3$  & AP$_4$  & AP$_r$  & AP$_c$  & AP$_f$  \\
\midrule
(1) & train & fc-cls & \emph{23.79} & 8.16 & 24.42 & 23.35 & 29.26 & 14.36 & 23.04 & 28.50 \\
(2) & train & head     & \emph{21.18} & 9.34 & 21.32 & 20.94 & 25.69 & 12.39 & 20.67 & 25.31 \\
\midrule
(3) & tune & head     & \emph{23.88} & 8.90 & 23.96 & 23.78 & 29.44 & 14.19 & 23.08 & 28.75 \\
(4) & tune & all      & \textbf{\emph{24.02}} & 8.91 & 24.86 & 23.49 & 29.06 & 14.81 & 23.36 & 28.52 \\
\bottomrule
\end{tabular}
\end{center}
\caption{Different ways to train models. Mode ``train" means train from random initialization. Mode ``tune" means finetune from trained model (1). Part \emph{fc-cls}, \emph{head}, and \emph{all} indicate the last classification FC layer, the whole classification head (2FC+fc-cls), and the whole backbone except for Conv1 and Conv2. $\beta$ is set to 1 here so that the results are lower than that in the main paper where $\beta=8$.}
\label{tab:frame}
\end{table}

There are several options to train a model with our proposed BAGS module.
As shown in Tab.\ref{tab:frame}, we try different settings with $\beta =1$. 
Since adding categories \emph{others} changes the dimension of classification outputs, we need to randomly initialize the classifier weights $W$ and bias $b$.
So for model (1), following~\cite{kang2019decoupling} to decouple feature learning and classifier, we fix all the parameters for feature extraction and only train the  classifier with parameters $W$ and $b$.
For model (2), we fix the backbone parameters and train the whole classification head together (2 FC and $W, b$).
It is worth noticing that the 2 FC layers are initialized by model (1), while $W, b$ are randomly initialized.
This drops mAP by 2.6\%, which may be caused by the inconsistent initialization of feature and classifier.
Therefore, we try to train $W$ and $b$ first with settings for model (1), and fine-tune the classification head (model (3)) and all backbones except for Conv1 and Conv2 (model (4)) respectively.
Fine-tuning improves mAP slightly.
However, taking the extra training time into consideration, we choose to take the setting of model (1) to directly train parameters for classifier only in all the other experiments.

\subsection{Comparison with winners of \href{https://www.lvisdataset.org/challenge}{LVIS 2019}}
Since the evaluation server for LVIS \emph{test} set is closed, all results in this paper are obtained on \emph{val} set.
There are two winners: \emph{lvlvisis} and \emph{strangeturtle}.
We compared  with \emph{lvlvisis} in Tab.3 based on their report~\cite{wang2019classification}, and our results surpass theirs largely.
For \emph{strangeturtle}, their Equalization Loss~\cite{tan2019equalization} (released on 12/11/2019) replaces   softmax with sigmoid for classification and blocks some back-propagation for tail classes.
With Mask R-CNN R50 baseline (mAP 20.68\%), Equalization Loss achieves 23.90\% with COCO pre-training (vs 26.25\% of ours). 
Our method performs much better on tail classes (AP$_r$ 11.70\%~\cite{tan2019equalization} vs 17.97\% ours).
They also tried to decrease the suppression effect from head over tail classes, but using sigmoid completely discards all suppression among categories, even though some of them are useful for suppressing false positives.
Without bells and whistles, our method outperforms both winners on \emph{val} set.

\subsection{Results on COCO-LT}

\begin{figure}
\begin{center}
	\includegraphics[width=\linewidth]{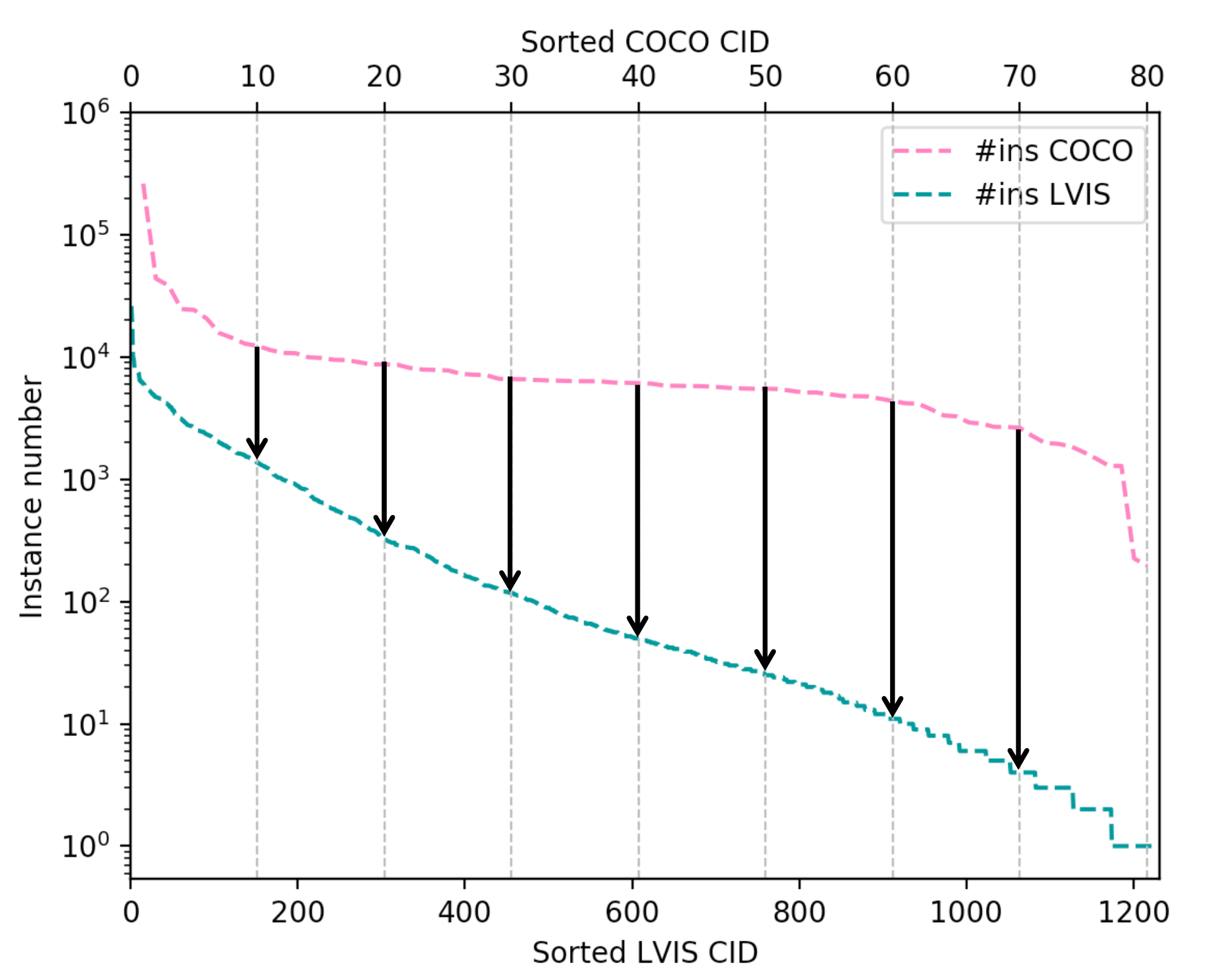}
	\caption{We align 80 categories of COCO with 1230 categories of LVIS, and sample corresponding number of instances for each COCO category.}
	\label{fig:show}
\end{center}
\end{figure}

To further verify the generalization ability of our BAGS, we construct a long-tail distribution dataset COCO-LT by sampling images and annotations from COCO~\cite{lin2014microsoft} train 2017 split.

\subsubsection{Dataset construction}
To get a similar long-tail data distribution as LVIS,
we first sort all categories of LVIS and COCO by their corresponding number of training instances.
As shown in Fig.~\ref{fig:show}, we  align 80 categories of COCO with 1230 categories of LVIS, and set the target training instance number  per category in COCO as the training instance number of its corresponding category in LVIS.
Then, we sample target number of instances for each COCO category.
We make use of as many instances in a sampled image as possible.
Training instances in a sampled image will only be ignored when there are plenty of instances belonging to that category.
In this way, we sample a subset of COCO that follows long-tail distribution just like LVIS.
COCO-LT only contains 9100 training images of 80 categories, which includes 64504 training instances.
For validation, we use the same validation set as COCO val 2017 split, which includes 5000 images.

\subsection{Main results}
We compare with Faster R-CNN and Mask R-CNN  (R50-FPN backbone) on the above COCO-LT dataset. The results are shown in Tab.~\ref{coco}.
Since the number of training images is small, we initialize baseline models with model trained on LVIS.
As we can see, our models  introduce more than 2\% improvements on mAP of both bounding box and mask. Importantly, it gains large improvement on tail classes.

\begin{table}[]
\small
\renewcommand{\tabcolsep}{1.8pt}
\renewcommand{\arraystretch}{1.3}
\begin{center}
\begin{tabular}{c|c|cccc}
\toprule
                & \textbf{mAP}  & {AP$_1$} & AP$_2$ & AP$_3$ & AP$_4$ \\
\midrule              
Faster R-CNN    & \emph{20.3} & {0.1}    & 12.9   & 24.3   & 26.7   \\
\textbf{Ours}            & \textbf{\emph{22.5}} & 13.0            & 18.6   & 24.1   & 26.4   \\
\midrule
Mask R-CNN bbox & \emph{19.1} & 0.0             & 11.1   & 22.9   & 26.4   \\
\textbf{Ours}            & \textbf{\emph{21.5}} & 13.4            & 17.7   & 22.5   & 26.0   \\
Mask R-CNN segm & \emph{18.0} & 0.0             & 11.5   & 21.8   & 23.3   \\
\textbf{Ours}            & \textbf{\emph{20.3}} & 3.4             & 18.9   & 21.7   & 23.0   \\
\bottomrule
\end{tabular}
\end{center}
\caption{Results on COCO-LT dataset. ResNet50-FPN backbone are used for both Faster R-CNN and Mask R-CNN.}
\label{coco}
\end{table}

{\small
\bibliographystyle{ieee_fullname}
\bibliography{egbib}
}

\end{document}